\documentclass[10pt,twocolumn,letterpaper]{article}

\usepackage{cvpr}              
\usepackage{multirow}  
\usepackage{svg}
\usepackage{siunitx}
\usepackage{graphicx}
\definecolor{cvprblue}{rgb}{0.21,0.49,0.74}
\usepackage[pagebackref,breaklinks,colorlinks,allcolors=cvprblue]{hyperref}

\def\paperID{*****} 
\def\confName{CVPR}
\def\confYear{2026}
\newcommand{\variable}[1]{\ensuremath{\boldsymbol{\mathrm{#1}}}}

\newcommand{\coordframe}[2]{\,^{\mathrm{#1}}\!\variable{T}_{\mathrm{#2}}}

\title{Light Field Based 6DoF Tracking of Previously Unobserved Objects}

\author{
    Nikolai Goncharov\thanks{\href{mailto:nikolai.goncharov@sydney.edu.au}{nikolai.goncharov@sydney.edu.au}} \qquad  James L. Gray \thanks{\href{mailto:james.gray1@sydney.edu.au}{james.gray1@sydney.edu.au}} \qquad Donald G. Dansereau\thanks{\href{mailto:donald.dansereau@sydney.edu.au}{donald.dansereau@sydney.edu.au}}\\\\University of Sydney\thanks{Australian Centre For Robotics (ACFR), School of Aerospace, Mechanical and Mechatronic Engineering, The University of Sydney, 2006 NSW, Australia.}
}

\begin{document}
\maketitle

\begin{abstract}
Object tracking is an important step in robotics and autonomous driving pipelines, which has to generalize to previously unseen and complex objects. Existing high-performing methods often rely on pre-captured object views to build explicit reference models, which restricts them to a fixed set of known objects. However, such reference models can struggle with visually complex appearance, reducing the quality of tracking. In this work, we introduce an object tracking method based on light field images that does not depend on a pre-trained model, while being robust to complex visual behavior, such as reflections. We extract semantic and geometric features from light field inputs using vision foundation models and convert them into view-dependent Gaussian splats. These splats serve as a unified object representation, supporting differentiable rendering and pose optimization. We further introduce a light field object tracking dataset containing challenging reflective objects with precise ground truth poses. Experiments demonstrate that our method is competitive with state-of-the-art model-based trackers in these difficult cases, paving the way toward universal object tracking in robotic systems. Code/data available at \color{red}{\href{}{https://github.com/nagonch/LiFT-6DoF}}. 
\end{abstract}
\section{Introduction}

\begin{figure}
    \centering
    \includegraphics[width=\linewidth]{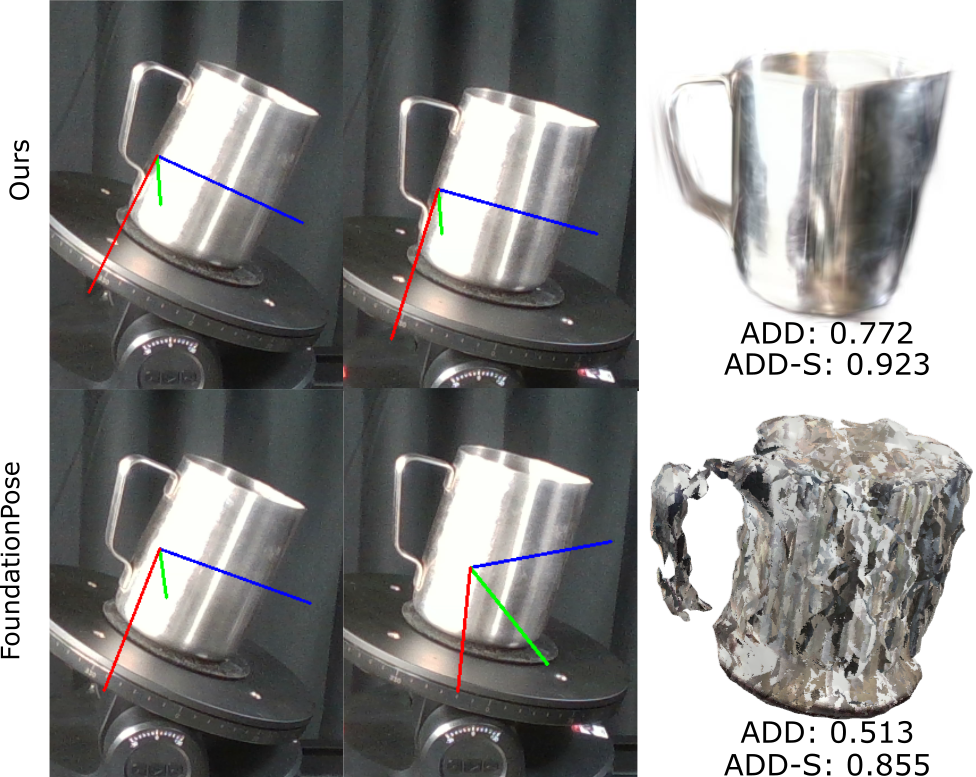}
    \caption{Our method compared with FoundationPose in the Jug (Tilt) sequence from our light field dataset. Our method not only outperforms FoundationPose in tracking the highly reflective object, it produces a high quality visual  reference model for a previously unseen object.}
    \label{fig:placeholder}
\end{figure}
Object tracking remains a crucial task for understanding motion in robotics and autonomous driving. In these settings, systems often encounter challenging factors such as previously unseen object categories and objects with shiny or reflective surfaces. Existing \textit{category-level} object tracking methods are either \textit{model-based} \cite{wang2019normalized, zhang2022ssp, zheng2023hs}, requiring a pre-defined CAD reference model, or \textit{model-free} \cite{park2020latentfusion, sun2022onepose, wen2024foundationpose}, relying on a pre-captured image set of the object to build the model offline. Moreover, these methods typically require RGB-D input and use mesh-based object representations, which are less effective for objects with complex visual appearance.

Recent advancements in vision foundation models \cite{kirillov2023segment, ravi2024sam, liu2024grounding, yang2024depth, yang2024depth2} enable the extraction of rich semantic and geometric features directly from images. Concurrently, neural scene representation methods \cite{mildenhall2021nerf, kerbl20233d} provide a robust way to model object appearance. Moreover, light field images \cite{bergen1991plenoptic, levoy2023light, ng2005light} capture the same plenoptic function approximated by those representations, while their dense multi-view sampling provides geometric cues \cite{wanner2013variational, jeon2015accurate, wang2022disentangling, chao2023learning} comparable to RGB-D sensing.

In this work, we leverage these advancements to move towards \textit{universal} object tracking. We introduce a light field image object tracking method that is robust to complex visual behavior without relying on a pre-trained reference model. We achieve this by integrating the semantic and geometric understanding provided by vision foundation models with appearance supervision from an online reference object model represented by Gaussian splats. We derive view-dependent Gaussian splats directly from light field observations, allowing for rendering-based pose optimization. Finally, for evaluation, we introduce a light field object tracking dataset featuring objects with highly specular and visually challenging surfaces and ground truth object and camera poses.

Our key insight is that light field images inherently serve as the very object models that prior methods must pre-build from reference images -- they simultaneously capture both geometry and appearance, providing a natural foundation for online pose tracking. To summarize, we make the following contributions:

\begin{itemize}
    \item
        A pose tracking method based on light field images that builds an object model online and without pre-recorded reference views, paving the way for universal object tracking.

    \item 
        Feed-forward conversion of light field images into view-dependent Gaussian splats, serving as an online tracking reference.

    \item
        A pose optimization module based on view-dependent Gaussian splats, leading to strong generalization for specular objects and performing on the level of state-of-the-art pose tracking methods.

    \item
        A light field pose tracking dataset with challenging specular objects, enabling rigorous evaluation of robustness and generalization.
        
\end{itemize}

These contributions advance toward open-world object tracking by removing dependence on pre-trained object models and handling objects with complex reflective behavior, which is essential for robotics and autonomous driving settings.

\section{Related Work} %
\textbf{CAD Model-based Object Pose Tracking}.
Classical approaches to the problem of object pose tracking typically require an \textit{instance-level} 3D CAD model of an object and focus on determining the pose of this object from a single image~\cite{lepetit2005monocular}. 
These classical approaches often used sparse features such as SIFT or SURF to establish correspondences and frame the pose tracking problem as the perspective-N-point problem~\cite{lepetit2005monocular, collet_moped_2011}.
Unfortunately, using sparse features discards significant amounts of information.
Accordingly, these methods perform poorly on textureless objects where features may not be present~\cite{li2018deepim}.
Learning-based approaches have subsequently improved upon these works~\cite{li2018deepim} and yielded state-of-the-art results~\cite{nguyen2025gotrack, wen2024foundationpose}.
However, this style of approach still requires an instance-level model of the object prior to tracking.

Some methods have attempted to relax the assumption of an instance-level model using a \textit{category-level} model instead.
These methods allow one to track any instance of an object within a known category~\cite{wang20206, lin2022keypoint}. 
Category-level object tracking is a more difficult problem, since the objects can vary in size and shape within the category.
Although these methods significantly broaden the range of objects that can be tracked from a single model, they still require an initial model.
On the other hand, our approach does not require the presence of an instance- or category-level model and can track previously unobserved objects.
\\
\textbf{Model-free Object Tracking.}
 Several machine-learning approaches removed the requirement of a CAD model of the object entirely and perform model-free tracking, instead relying on a series of reference images of an object to pre-train a model for the object prior to tracking. 
Using reference images instead of a CAD model significantly lowers the barrier to entry for object tracking, and these methods achieve impressive results~\cite{park2020latentfusion, wen2024foundationpose}. 
However, these approaches still require foreknowledge of the object to track and 
in this work, we remove the requirement of foreknowledge of the object. 

Some approaches, such as BundleTrack and BundleSDF, attempt to perform object tracking without prior knowledge of the object~\cite{wen_bundletrack_2021, wen2023bundlesdf}. 
Instead of a series of reference RGB-D images, these methods simply use image segmentation in the first frame of a video sequence to specify the object.
Note that these methods do not perform as well as methods that use reference images, such as FoundationPose~\cite{wen2024foundationpose}.
Furthermore, these approaches rely on the use of an RGB-D camera, which has been shown to face significant challenges in estimating depth in visually challenging scenes~\cite{RGB_D_Cai}. 
\\ %
\textbf{Light Field Imaging and Object Tracking}. 
A traditional photograph captures a 2D 
projection of the 5D plenoptic function.
In contrast, light field imaging captures a 4D slice of the plenoptic function
and retains the angular information lost by conventional
photography~\cite{wu_light_2017}. In fact, a light field can fully represent the
5D plenoptic function, since each ray in the plenoptic function retains its
brightness in the direction of travel~\cite{levoy_light_2006}. Light field image
processing methods have been used to perform several tasks such as depth
estimation, scene-flow~\cite{ma_differential_2020, gray2025gradient}, deblurring, view synthesis, super-resolution~\cite{wu_light_2017,
zhou_review_2021}, and object 
tracking~\cite{wang_light_2026, wang2023_object}. 
Most recent light field object tracking methods do not perform 6DoF object pose tracking and  instead compute a bounding box for the objects.
On the other hand, our approach performs 6DoF pose tracking.

In this paper, we use light field depth estimation as an alternative to RGB-D cameras in visually challenging environments.
There have been many approaches to light field depth estimation.
Traditional approaches have taken many forms including global variational approaches that resemble traditional optical flow and stereo approaches~\cite{tran_variational_2017, gray2025multi}, depth from defocus~\cite{tao2013depth} and epipolar plane image (EPI) based methods that rely on finding slopes in the EPI~\cite{wanner2013variational, wu_light_2017}.
Machine learning has also been applied to light field depth estimation and yielded state-of-the-art results~\cite{OccCasNet_2024, Gao_2025_ICCV}.
\\ %
\textbf{Radiance Fields}.
In the original NeRF, a fully-connected deep network was used to map spatial locations and viewing directions to an emitted radiance and volume density~\cite{mildenhall2021nerf}. 
Although NeRFs perform high-quality novel view synthesis, even in the presence of specular objects~\cite{verbin_ref-nerf_2021, wang_unisdf_2024}, these methods are implicit.
This means that it is difficult to compare two NeRFs without comparing the renders. 
On the other hand, Gaussian Splatting represents scenes as collections of 3D Gaussians, each with a position, scale, opacity, and view-dependent color, Gaussian splatting offers real-time renders of quality comparable to that of NeRFs~\cite{kerbl20233d}.
Gaussian splatting based approaches have also been shown to accurately model the appearance of highly specular objects~\cite{ye_3d_2024, gu_irgs_2025}.

Radiance field approaches have also been applied to object tracking.
Some of these approaches formulate pose-tracking as a SfM localization or SLAM problem. 
After an initial coarse estimate of the geometry of the scene, joint reconstruction and pose optimization is performed~\cite{wen2023bundlesdf, chen_gsgtrack_2024, jin_6dope-gs_2025}.
These methods lend themselves well to modeling the appearance of objects and their geometry.
We have chosen to use to model the appearance of objects using Gaussian splats because they form an explicit representation of the object.
\\
\textbf{Foundation Models}
Once trained, foundation models can be used as-is, adapted or fine-tuned for use in down-stream tasks.
These models typically demonstrate excellent performance owing to the scale and diversity of the datasets used to train the and their size~\cite{awais_foundation_2025}, e.g. Grounding Dino has 172M parameters~\cite{liu2024grounding}, Depth Anything v2 has 335M~\cite{yang2024depth}.
Often, the behaviour of these models can be modified using prompts, such as a bounding box or a point to indicate which item to segment in the case of SAM2~\cite{ravi2024sam}.
In this work, we use Grounding DINO and SAM2 in our segmentation stage (see Section~\ref{subsec:LF_Processing}), and we build on the work of Video Depth Anything to obtain accurate and robust metric accurate depth maps for each frame of light field video~\cite{Chen_2025_CVPR}. 

\section{Method}

\begin{figure*}[t]
    \centering
    \includegraphics[width=\linewidth]{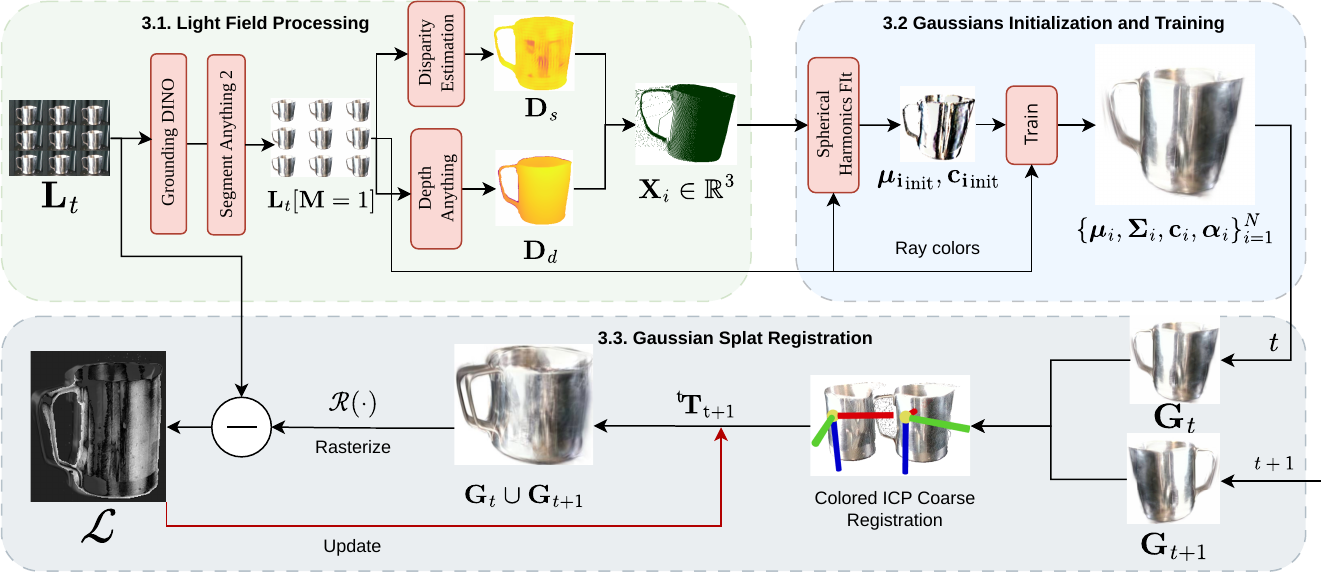}
    \captionof{figure}{An overview of our method of light field based object tracking. Given a light field image $\variable{L}_t$ of a previously unseen object, we obtain an object mask $\variable{M}$ using Grounded DINO \cite{liu2024grounding} and SAM 2 \cite{ravi2024sam}. The light field is then converted into a sparse disparity map $\variable{D}_s$, fused with dense disparity $\variable{D}_d$ from Depth Anything \cite{yang2024depth}, and unprojected to produce the object point cloud $\variable{X}$  (Sec.~\ref{subsec:LF_Processing}). Using $\variable{X}$ and  rays, we initialize a Gaussian object representation with centroids $\variable{\mu}_i$ and view-dependent colors $\variable{c}_i$, then train it to obtain the Gaussian of the frame, $\variable{G}_t$ (Sec.~\ref{subsec:gs_init}). To register $\variable{G}_t$ and $\variable{G}_{t+1}$, we first run Colored ICP \cite{park2017coloredicp} on centroids and colors to get a coarse transform $\coordframe{t}{t+1}$, then refine it by jointly rasterizing the concatenated Gaussians under supervision from $\variable{L}_t$ (Sec.~\ref{subsec:gs_reg}).}
    \label{fig:pipeline}
\end{figure*}

Let $\coordframe{}{w}$ denote the pose of a static calibrated light field camera observing an object of interest across two light field frames: $\variable{L}_t$ and $\variable{L}_{t+1}$. The corresponding object poses, $\coordframe{w}{t}$, and $\coordframe{w}{t+1}$, are unknown. The goal is to estimate the relative transform $\coordframe{t}{t+1}$ at time $t$ and $t+1$.

An overview of our method is shown in \autoref{fig:pipeline}. We first \textit{segment} the object across all light field views using Grounding DINO \cite{liu2024grounding} and SAM2 \cite{ravi2024sam}, then \textit{estimate depth} by fusing disparities produced by light field processing and Depth Anything Video \cite{Chen_2025_CVPR}, and finally \textit{unproject} the fused disparity to obtain the object’s 3D point cloud (\autoref{subsec:LF_Processing}). We then \textit{initialise Gaussian splats} by fitting spherical harmonic coefficients to the view-dependent color observations across subviews (\autoref{subsec:gs_init}). These Gaussians then \textit{coarsely registered} using Colored ICP \cite{park2017coloredicp} and \textit{refined} through photometric optimization on the light field subviews (\autoref{subsec:gs_reg}). As a result, we estimate the object’s relative motion $\coordframe{t}{t+1}$ by optimizing the transformation that best aligns the Gaussian splats between consecutive frames.

In our method, we express light field rendering in the form of Gaussian splatting, enabling joint rendering across multiple frames by concatenating their 3D Gaussians, which for irregular motion is challenging in grid-based light field rendering. This preserves the multi-view benefits of light fields while simplacknowlifying motion estimation in 3D space, making the Gaussians our online tracking reference model.

\subsection{Light Field Processing}\label{subsec:LF_Processing}
\textbf{Segmentation}  \quad
    In a setting similar to \cite{goncharov2025segment}, given a 4D light field image $\variable{L}$, we aim to segment an object of interest, described by a text prompt (e.g., ``jug"). We first use Grounding DINO \cite{liu2024grounding} to detect the object in a chosen reference subview, obtaining a bounding box. The bounding box is then used to prompt SAM2 \cite{ravi2024sam}, which segments the object and tracks it across all subviews, yielding a binary-valued 4D light field object mask $\variable{M}$. We continue our work with the light field  $\variable{L}[\variable{M} = 1]$.
\\ %
\\
\textbf{Point Cloud Extraction} \quad 
We aim to extract a dense point cloud from the light field with sufficient geometric precision to support accurate registration across views. We first process the light field into a \textit{sparse disparity map} $\variable{D}_s \in \mathbb{R}^{U \times V}$ using the structure tensor formulation of Goldluecke and Wanner \cite{wanner2013variational} and taking values with high confidence $\variable{C} > \variable{C}_{\mathrm{thresh}}$.

Then, in order to benefit from powerful depth priors, we apply Depth Anything \cite{yang2024depth} to the central subview $\variable{L}[s_m, t_m]$, obtaining a \textit{dense relative disparity map} $\variable{D}_d$. The geometry of $\variable{D}_s$ is then used to align the relative map by fitting a linear transformation $\variable{D}_s = \alpha \cdot \variable{D}_d + \beta$, where $\alpha$ and $\beta$ are obtained through least-squares regression.
Assuming the subviews are undistorted, share the same previously estimated intrinsics $\variable{K}$, and adjacent views are separated by a fixed baseline $b$, we unproject each middle view pixel $(u, v)$ using its disparity to get the resulting point cloud $\variable{X} \in \mathbb{R}^3$:
\begin{equation}
    \variable{X}(u, v) = \left( \frac{f \cdot b}{\alpha \cdot \variable{D}_d(u, v) + \beta} \right) \variable{K}^{-1} \begin{bmatrix}
        u \\ v \\ 1
    \end{bmatrix}.
\end{equation}

\subsection{Gaussian Splatting Initialization and Training}\label{subsec:gs_init}
\textbf{Initialization} \quad
In order to initialize per-point spherical harmonic coefficients, we first obtain the color values of the point in all the light field subviews. Given a reconstructed point cloud $\variable{X}$ and a light field, we associate each point  with the set of its corresponding light field rays. The projection $\variable{\Pi} (\variable{X_i})$ uses 
light field camera intrinsics and object mask $\variable{M}$ to map each point to a set of RGB color 
samples across all subviews, yielding $\variable{\Pi} : \mathbb{R}^3 \rightarrow \mathbb{R}^{k \times 3}$, 
where the number of rays $k$ depends on a particular point's visibility.

For each point $\variable{X_i}$, given its set of rays $\variable{\Pi} (\variable{X_i})$, we estimate the per-point spherical harmonic coefficients by solving a weighted least-squares problem, following the formulation in Wiersma et al. \cite{wiersma2025uncertainty}. Specifically, we fit the coefficients $\variable{c_i}$ such that the spherical harmonic basis functions $\variable{Y}$ best approximate the observed radiance values:
\begin{equation}
    (\variable{Y}^T \variable{Y} + \lambda \variable{W}) \variable{c}_i = \variable{Y}^T \variable{\Pi} (\variable{X_i}).
\end{equation}
Here, $\variable{Y}$ is the matrix of spherical harmonic basis functions evaluated at the viewing directions corresponding to the observed rays of $\variable{X_i}$, $\lambda$ is a regularization coefficient, and $\variable{W} = e^l\variable{I}$ is the diagonal regularization term that penalizes higher-order coefficients. In our implementation, we use harmonics up to degree $l = 2$. We use a constant initial opacity for all points and assign random orientations to numerically ensure proper initial optimization. To initialize the scale of the Gaussian, we back-project the pixel sizes through the camera model at the corresponding depth. As a result, we obtain an initialized set of Gaussians  ${\variable{G}} = \{{\variable{\mu}_i, \variable{\Sigma}_i, \variable{c}_i, \variable{\alpha}_i}\}_{i=1}^N$.
\\
\\
\textbf{Training of the Gaussians} \quad
To refine the appearance and spatial extent of each Gaussian corresponding to the light field image, we perform a brief online optimization stage following the standard Gaussian splatting procedure \cite{kerbl20233d}. Each light field subview supervises the reconstruction of its corresponding Gaussian set via the photometric objective:
\begin{equation}
\mathcal{L}(\variable{G}) = \sum_{i} \big| \mathcal{R}_i(\variable{G}) - \variable{I}_i \big|_1,
\end{equation}
where $\mathcal{R}_i$ is the function that rasterizes the Gaussians into subview $i$. Additionally, a mask-aware term encourages the rendered opacity to match the mask $\variable{M}$, ensuring accurate object boundaries and spatial consistency. The loss function is optimized using Adam \cite{kingma2014adam}.

This stage fine-tunes the initialized Gaussians $\variable{G}$ to better reproduce the observed light field appearance, improving both spatial precision and radiance fidelity without re-learning the underlying geometry.

\subsection{Gaussian Splat Registration}\label{subsec:gs_reg}
Given two time frames $\variable{t}$ and $\variable{t+1}$, two light field images $\variable{L}_t$ and $\variable{L}_{t+1}$, corresponding unknown object poses $\coordframe{w}{t}$ and $\coordframe{w}{t+1}$, and corresponding Gaussian splat sets $\variable{G}_t$ and $\variable{G}_{t+1}$, the task is to find relative object transform $\coordframe{t}{t+1}$. The camera remains static throughout the sequence, and therefore its pose in both frames is given by $\coordframe{}{w}$.
\\
\\
\textbf{Coarse Registration} \quad
We treat the Gaussian centers $\variable{\mu}_i$ as 3D point coordinates and the ambient color components $\variable{c}_i{_0}$ as corresponding color features. We first perform global alignment by voxel downsampling each set and computing fast point feature histograms \cite{rusu2009fast} on the downsampled points. Next, a RANSAC-based feature matching estimates a coarse initial pose transform that ensures global geometric consistency.

We then apply photometric–geometric refinement using the Colored ICP algorithm \cite{park2017coloredicp}, which jointly minimizes geometric and color residuals, obtaining a coarse relative transform:
\begin{align}
E(\variable{T}) =
&\sum_i 
\lambda_g \, \| \mathbf{T} \{\variable{\mu_i}\}_t - \{\variable{\mu_i}\}_{t+1} \|^2 \nonumber \\
&+ \lambda_c \, \| \{\variable{c}_i{_0}\}_{t} - \{\variable{c}_i{_0}\}_{t+1} \|^2,
\label{eq:colored_icp}
\end{align}
where $\lambda_g$ and $\lambda_c$ weight geometric and photometric terms.
\\
\\
\textbf{Pose Refinement} \quad
We start with a coarse inter-frame transform ${\,^{\mathrm{t}}\!\variable{\tilde{T}}_{\mathrm{t+1}}}$ estimated during the previous step. Our objective is to refine this transform by optimizing for an estimate that aligns the two Gaussian representations $\variable{G}_t$ and $\variable{G}_{t+1}$ and minimizes the photometric loss in the light field subviews.
\\
We optimize for the $\mathrm{SE(6)}$ form of the relative transform $\variable{\xi} = \mathrm{log}_{\mathrm{SE(3)}}(\variable{T})$. At each step, we transform the Gaussian $\variable{G}_{t+1}$ into the frame $t$ by applying:
\begin{align}
\label{eq:mean-xform}
\boldsymbol{\mu}'_i \;&=\; {\mathbf{R}}\,\boldsymbol{\mu}_i + \mathbf{t},\\
\label{eq:cov-xform}
\boldsymbol{\Sigma}'_i \;&=\; \mathbf{R}\,\boldsymbol{\Sigma}_i\,\mathbf{R}^{\!\top},\\
\label{eq:sh-xform}
\mathbf{c}'_{i,\ell} \;&=\; \mathbf{D}_\ell\!\big(\mathbf{R}\big)\,\mathbf{c}_{i,\ell},
\end{align}
where $\mathbf{D}_\ell(\cdot)$ is the Wigner D-matrix of spherical harmonics rotation for degree $\ell$. Then, the transformed Gaussian is $\variable{G}_{t+1}(\variable{\xi})' = \{\{ \variable{\mu}_i', \variable{\Sigma}_i', \variable{c}_{i, \ell}', \variable{\alpha}_i\}\}$.

\begin{equation}
\mathcal{L} = \sum_{i}\, \big\| \mathcal{R}_i(\variable{G}_t \cup  \variable{G}_{t+1}(\variable{\xi})') - \variable{I}_i \big\|_1.
\end{equation}
\\
This joint optimization across subviews enforces photometric consistency between frames and yields a refined inter-frame transform.

\section{Experiments}
\begin{table*}[t]
\centering
\caption{Quantitative results for the model-free mode of FoundationPose \cite{wen2024foundationpose} vs our method on the captured sequences. Our proposed method demonstrates performance competitive with FoundationPose, discussed in Sec.~\ref{subsec:quant_results}.}
\begin{tabular}{c c c c c c c c }
\toprule
Sequence & Method & ADD $\uparrow$ & ADD-S $\uparrow$ &
$\Delta R_{abs}$ (°) $\downarrow$ &
$\Delta t_{abs}$ (m) $\downarrow$ &
$\Delta R_{rel}$ (°) $\downarrow$ &
$\Delta t_{rel}$ (m) $\downarrow$ \\
\midrule
\midrule

\multirow{2}{*}{Jug} & Foundation Pose
 &\textbf{ 0.909} & \textbf{0.961} & \textbf{7.061} & \textbf{7.084e-3} & 5.520 & \textbf{3.338e-3} \\
 & Ours  & {0.805} & 0.941 & 26.360 & 1.018e-2 & \textbf{4.638} & 7.373e-3 \\
\midrule

\multirow{2}{*}{Jug (tilt)} & FoundationPose
 & 0.513 & 0.855 & 53.812 & 3.060e-2 & 11.027 & 1.645e-2 \\
& Ours 
 & \textbf{0.772} & \textbf{0.923} & \textbf{8.304} & \textbf{2.134e-2} & \textbf{5.513} & \textbf{8.311e-3} \\
\midrule

\multirow{2}{*}{Jug ($Z$ Motion)} & FoundationPose
 & \textbf{0.714} & \textbf{0.937} & \textbf{42.933} & 9.755e-3 & 7.822 & \textbf{8.065e-3} \\
& Ours
 & 0.229 & 0.817 & 45.284 & \textbf{3.796e-2} & \textbf{7.821} & 1.242e-2 \\
\midrule

\multirow{2}{*}{Tea Box} & FoundationPose
 & \textbf{0.932} & \textbf{0.967} & \textbf{5.240} & \textbf{5.787e-3} & \textbf{2.603} & \textbf{2.015e-3} \\
& Ours
 & 0.702 & 0.913 & 29.252 & 8.783e-3 & 2.705 & 5.511e-3 \\
\midrule

\multirow{2}{*}{Tea Box (tilt)} & FoundationPose
 & 0.626 & 0.855 & \textbf{5.692} & 3.736e-2 & \textbf{1.602} & 6.591e-3 \\
& Ours
 & \textbf{0.759} & \textbf{0.942} & 8.965 & \textbf{2.233e-2} & 2.835 & \textbf{5.317e-3} \\
\midrule

\multirow{2}{*}{Box} & FoundationPose
 & 0.859 & 0.938 & \textbf{1.473} & 1.321e-2 & \textbf{0.974} & \textbf{2.462e-3} \\
& Ours
 & \textbf{0.891} & \textbf{0.950} & 2.067 & \textbf{1.041e-2} & 1.202 & 2.842e-3 \\
\midrule

\multirow{2}{*}{Shiny Box} & FoundationPose
 & 0.824 & 0.947 & \textbf{8.583} & 1.579e-2 & 11.027 & 1.645e-2 \\
& Ours
 & \textbf{0.850} & \textbf{0.958} & 9.357 & \textbf{1.118e-2} & \textbf{1.731} & \textbf{5.805e-3} \\
\midrule
\midrule
\multirow{2}{*}{Average} & FoundationPose
 & \textbf{0.768} & 0.923 & \textbf{17.828} & \textbf{1.708e-2} & 4.632 & \textbf{6.345e-3} \\
& Ours
 & 0.755 & \textbf{0.928} & 18.513 & 1.745e-2 & \textbf{3.778} & 6.797e-3 \\
\bottomrule
\end{tabular}
\label{tab:pose_tracking_results}
\end{table*}

\begin{table*}[t]
\centering
\caption{Quantitative results for the model-free mode of FoundationPose \cite{wen2024foundationpose} vs our method on our synthetic sequences. On the synthetic sequences both methods used ground-truth depths instead of estimated depths. We note that the performance of our method does not change significantly when the car is made shiny.}
\begin{tabular}{c c c c c c c c}
\toprule
Sequence & Method  & ADD $\uparrow$ & ADD-S $\uparrow$ &
$\Delta R_{abs}$ (°) $\downarrow$ &
$\Delta t_{abs}$ (m) $\downarrow$ &
$\Delta R_{rel}$ (°) $\downarrow$ &
$\Delta t_{rel}$ (m) $\downarrow$ \\
\midrule

\multirow{2}{*}{Car} & FoundationPose
 & \textbf{0.747} & \textbf{0.877} & \textbf{8.374} & \textbf{2.948e-2} & \textbf{13.450} & \textbf{4.613e-2} \\
 & Ours
 & 0.149 & 0.442 & 53.736 & 1.165e-1 & 20.370 & 8.141e-2 \\
\midrule

\multirow{2}{*}{Shiny Car} & FoundationPose
 & \textbf{0.142} & \textbf{0.294} & 64.341 & \textbf{5.328e-2} & \textbf{28.333} & 8.890e-2 \\
 & Ours
 & 0.111 & 0.232 & \textbf{41.722} & 8.494e-2 & 28.398 & \textbf{8.227e-2} \\
\bottomrule
\end{tabular}
\end{table*}

\begin{figure*}[t]
    \centering
    \includegraphics[width=\linewidth]{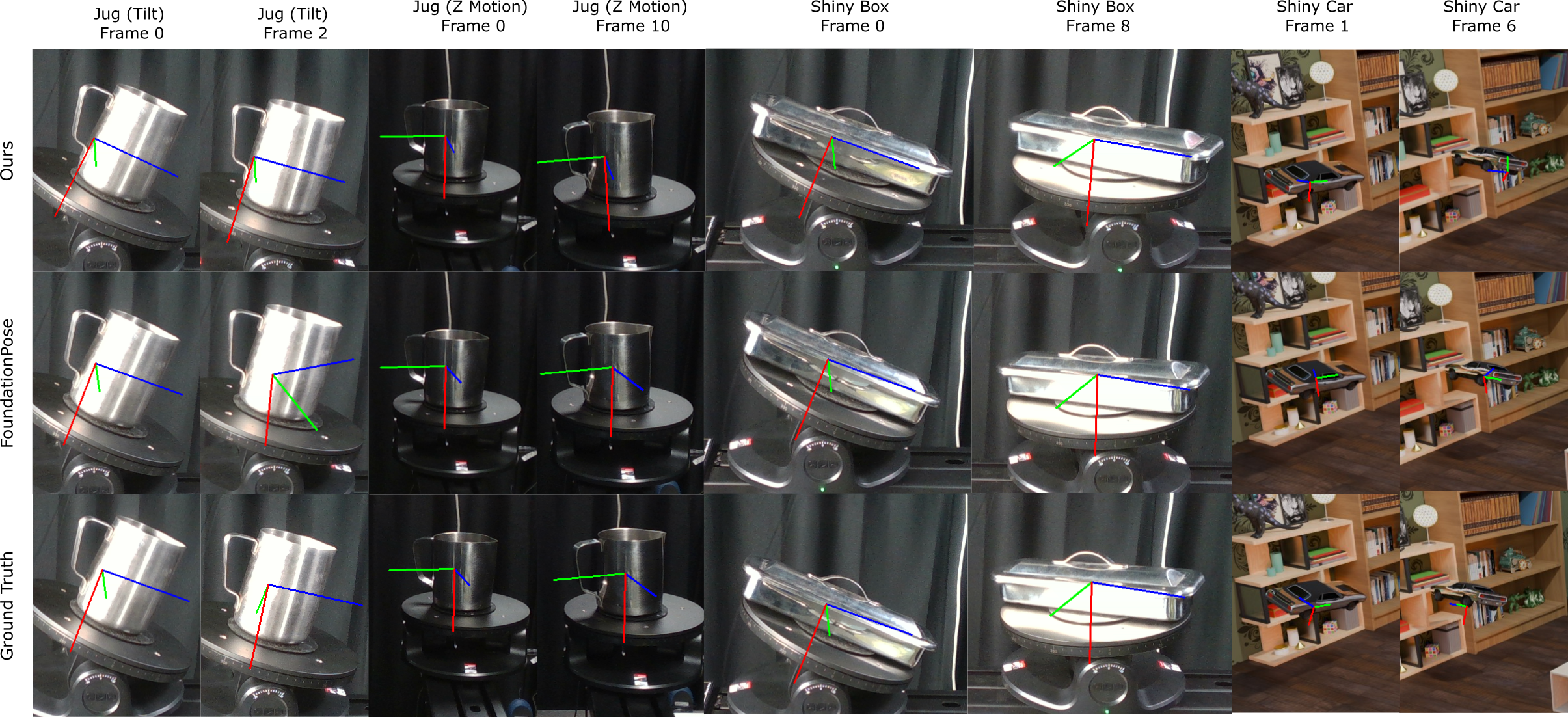}
    \caption{Qualitative comparison of our method and FoundationPose \cite{wen2024foundationpose}. On the Jug (Tilt) sequence FoundationPose exhibits noticeable drift as the tilt increases, whereas our method demonstrates increased robustness to viewpoint and lighting changes. On the Jug ($Z$ motion) our approach slightly underestimates the rotational magnitude, although the overall pose remains consistent. In the other sequences we demonstrate comparable performance with FoundationPose.}
    \label{fig:qual_results}
\end{figure*}

\begin{table*}[t]
\centering
\caption{Ablation studies of our method. Removing online Gaussian training weakens both rotation and translation, showing the role of appearance cues during refinement. Dropping Spherical Harmonics initialization hurts rotation most and increases translation error, since SH (Spherical Harmonics) features act as a stabilizing prior for Gaussian updates. Using ICP alone gives only moderate accuracy: reliable for short ranges, but prone to eventual drift.}
\label{table:ablation}
\begin{tabular}{ccccccc}
\toprule
Inference Mode &  ADD $\uparrow$ &  ADD-S $\uparrow$ & $\Delta R_{abs}$ (°) $\downarrow$ & $\Delta t_{abs}$ (m) $\downarrow$  & $\Delta R_{rel}$ (°) $\downarrow$  &  $\Delta t_{rel}$ (m) $\downarrow$\\
\midrule
\textbf{Full Method} & \textbf{0.772} & \textbf{0.923} & \textbf{ 8.304} & \textbf{0.021} & 5.51 & 8.04e-3\\
No Online Training & \textbf{\textcolor{red}{0.625}} & \textbf{\textcolor{red}{0.882}} & 18.04 & \textbf{\textcolor{red}{0.035}} & \textbf{\textcolor{red}{10.31}} & 8.34e-3\\
No SH Initialization & 0.648 & 0.902 & \textbf{\textcolor{red}{31.79}} & 0.029 & 4.89 & \textbf{\textcolor{red}{ 9.00e-3}}\\
ICP only & 0.734 & 0.917 & 15.42 & 0.022 & \textbf{4.43} & \textbf{7.00e-3} \\
\bottomrule
\end{tabular}
\label{tab:ablation2}
\end{table*}

\subsection{Dataset}
We evaluate our approach on a novel light field pose tracking dataset. This dataset consists of both natural and synthetic light field images of objects of varying specularities under a diversity of illumination to replicate realistic scenarios. Each light field image consists of a $9 \times 9$ subview grid with a 5 mm distance between the views. For all the natural and synthetic objects, we additionally provide 15 to 20 reference views for compatibility with FoundationPose. Each motion sequence is from 15 to 20 frames long.
\\
\textbf{Captured Sequences} \quad
This set of sequences was recorded in a controlled lab setup. Objects with diverse reflectivities and geometries (e.g., diffuse box, reflective jug) were mounted on motorized rotation and translation stages to obtain precise ground-truth 6D poses. The motion is predominantly horizontal, combined with controlled in-plane rotation and tilt. In practice, to obtain accurate depth supervision, we mount a RealSense depth camera on a robotic arm, which then traverses all subview positions and records both RGB and depth at each location. This setup ensures a fixed 5 mm baseline and provides reliable depth for the central view. The dataset includes full light field calibration, camera intrinsics, depth for the middle subview, and per-frame ground-truth object poses.
\\
\textbf{Synthetic Sequences} \quad
We generated the dataset in Blender \cite{Blender}.
Two synthetic sequences were rendered. Each sequence shows the same object (a toy car) undergoing the same projectile motion.
Both sequences include a cluttered 
background to introduce realistic distractors without compromising control over geometry, lighting, and ground truth. 
The only change between the sequences is surface reflectivity.
This allows us to directly determine the impact of surface reflectivity on our approach.

\subsection{Metrics}
Similar to \cite{xiang_posecnn_2017} and \cite{wen2024foundationpose} we use area under the curve (AUC) of ADD and ADD-S, using the threshold of 0.1. Additionally, we report a set of standard object tracking metrics:
\\
\textbf{Absolute pose accuracy}
\begin{itemize}
    \item Mean angular rotation error between prediction and ground truth, $\Delta R_{abs}$.
    \item Mean Euclidean translation error between prediction and ground truth, $\Delta t_{abs}$.
\end{itemize}
\textbf{Relative pose accuracy}
\begin{itemize}
    \item Mean relative rotation error $\Delta R_{rel}$, computed from predicted vs.\ ground-truth frame-to-frame motion.
    \item Mean relative translation error $\Delta t_{rel}$, capturing temporal consistency across consecutive frames.
\end{itemize}

\subsection{Experimental Setup}
We use FoundationPose \cite{wen2024foundationpose}, equipped with a pre-trained reference model, as a baseline for our evaluation. We evaluate both methods across all sequences in our dataset. For FoundationPose, we first reconstruct the object mesh from the designated reference views and then apply the method to all sequences. To isolate tracking performance, every sequence is initialized with the ground-truth pose of the first frame; no initial pose estimation is used.


After initialization, the Gaussians in each frame are optimized using Adam (learning rate 1e-3, 1000 steps). All remaining hyperparameters follow the standard configuration used in the GSplat training routine \cite{ye2025gsplat}. After registration, we refine the pose using the same optimizer and base learning rate for 400 iterations, decaying the learning rate by a factor of 0.1 every 20 iterations. In practice, we find that during this step, keeping colors, opacities and covariances optimized yields the most stable results. Additionally, while refining the poses, we only render for ambient color. For both methods, the mesh produced by FoundationPose is used when computing ADD and ADD-S.

Since Blender scenes contain ground truth depth maps of higher fidelity than ones produced by an RGB-D camera, we use them directly for this dataset in both FoundationPose and our method to ensure fair comparison.

\subsection{Quantitative Results} \label{subsec:quant_results}
\textbf{Captured Data} \quad
The qualitative estimation for our method is presented in Table~\ref{tab:pose_tracking_results}. On the captured dataset, absolute rotation error for our method is slightly higher on average than FoundationPose, but the gap reverses on reflective sequences: in the jug-tilt case FoundationPose exhibits large drift due to reflectivity, while our drift increases by only about 1°, due to pose integration. Absolute translation shows two regimes: on difficult objects such as the jug, our depth priors and light-field disparity outperform the depth-camera + mesh setup used by FoundationPose; on visually simple sequences with clean Z-motion, FoundationPose’s geometry priors give lower error. Averaged across all sequences, both methods converge to approximately 1.7 cm translation error. For relative metrics, our method is consistently stronger: relative rotation improves by about 1° on average, particularly on shiny objects, and relative translation stays matched at roughly 6 mm. ADD is marginally worse for our method because the evaluation of this metric relies on the mesh produced by FoundationPose, while ADD-S shows no meaningful difference. Overall, both systems achieve similar accuracy, with our method providing better stability on reflective objects and more reliable relative pose estimation.
\\
\textbf{Synthetic Data} \quad
On the fast-motion Car sequence, FoundationPose attains ADD value of 0.747, whereas our method reaches 0.149, indicating that rapid inter-frame motion undermines our ICP-based initialization. Under increased surface reflectivity (Shiny Car), FoundationPose drops from 0.747 to 0.142, while our performance remains largely unchanged. Overall, this shows our method's reduced robustness to abrupt motion and stability under challenging lighting and specular conditions.

\subsection{Qualitative Results}
Figure \ref{fig:qual_results} presents qualitative comparisons between our approach and FoundationPose. On the tilting jug sequence under strong illumination, FoundationPose exhibits noticeable drift as the tilt increases, whereas our method maintains stable tracking throughout the motion, demonstrating increased robustness to viewpoint and lighting changes. On the jug rotation around the Z-axis, our approach shows a mild tendency to underestimate the rotational magnitude, although the overall pose remains consistent. For the shiny box, although the object is visually challenging due to its reflectance, both methods maintain stable tracking. Our approach handles the reflections reliably, while FoundationPose also succeeds, likely aided by the object’s simple geometry. For the synthetic flying car sequence, we can see that the object's position gets reliably tracked across the multiple initial frames, but the rotation quickly diverges.

\begin{figure}
    \centering
    \includegraphics[width=\linewidth]{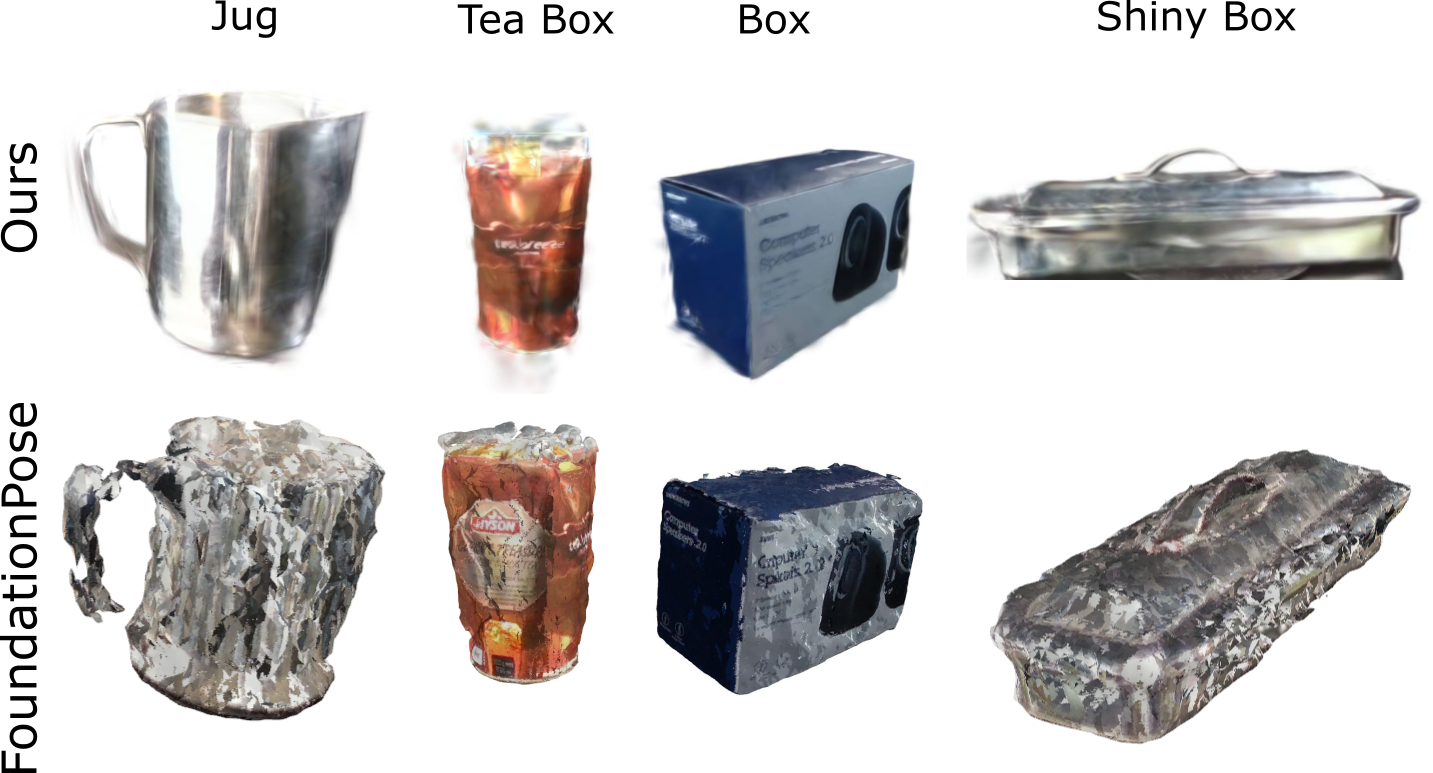}
    \caption{A visual comparison of FoundationPose's meshes that it uses for object-tracking and renders from our Gaussian Splats. Our method produces more photorealistic reference models.}
    \label{fig:ref_models}
\end{figure}

Additionally, Figure \ref{fig:ref_models} shows the visual differences between the meshes pre-captured by FoundationPose and our reference Gaussians. Due to the fact that RGB-D cameras assume Lambertian surfaces, the mesh reconstructions for the specular objects experience surface artifacts. However, our Gaussians not only show high fidelity appearance of the object, but also lock to the present lighting conditions.

\subsection{Ablation Studies}
Table~\ref{table:ablation} presents ablation studies for our method. We evaluate four inference modes: \textbf{Full Method}, which includes SH-based Gaussian initialization, online Gaussian optimization, and ICP; \textbf{No Online Training}, which initializes Gaussians once and aligns them without any parameter updates; \textbf{No SH (Spherical Harmonics) Initialization}, which starts Gaussian training from colorless points; and \textbf{ICP Only}, which bypasses Gaussians entirely and aligns raw geometry with color. The full pipeline attains the highest accuracy, serving as the performance ceiling. Removing online training worsens both rotation and translation performance, showing that appearance of the Gaussians during pose refinement is essential. Removing SH initialization degrades rotation the most and increases relative translation error, indicating that SH features provide a stabilizing prior for Gaussian parameters' optimization. Finally, ICP alone reaches moderate accuracy with good short-range alignment but predictable drift, reflecting the limitations of relying solely on geometry and color.

\section{Conclusion}
This work presents an object-tracking framework that removes the dependence on pre-trained reference models by exploiting the richer geometric cues available in light field images. By treating the light field itself as the reference, the method can track previously unseen objects while maintaining stable performance. Robust initialization and a Gaussian-splatting refinement stage allow the algorithm to recover accurate relative poses for visually challenging objects. To support progress in this direction, we also provide the first dedicated dataset for light-field-based object tracking, with ground-truth annotations for benchmarking.

Across a range of difficult scenarios that reflect real robotic operation, the approach reaches the performance level of FoundationPose while requiring no prior object information. The main drawback lies in the computational load of per-frame light field processing and the absence of global trajectory optimization. A natural next step is to exploit multi-frame consistency and accumulate splats across time, enabling joint tracking and reconstruction. This would move light-field-based tracking closer to practical deployment in robotics and autonomous systems.

\section{Acknowledgements}
This research was supported in part by funding from Ford Motor Company, the Australian Research Council Research Hub in Intelligent Robotic Systems for Real-Time Asset Management (ARIAM, IH210100030), and NVIDIA Academic Grant Program.

{
    \small
    \bibliographystyle{ieeenat_fullname}
    \bibliography{main}

@String(CVPR= {IEEE Conf. Comput. Vis. Pattern Recog.})

@String(ICCV= {Int. Conf. Comput. Vis.})

@String(ECCV= {Eur. Conf. Comput. Vis.})

@String(ICME = {Int. Conf. Multimedia and Expo})

@String(CVPRW= {IEEE Conf. Comput. Vis. Pattern Recog. Worksh.})

@String(CVPR  = {CVPR})

@String(ICCV  = {ICCV})

@String(ECCV  = {ECCV})

@String(ICME  =	{ICME})

@String(CVPRW= {CVPRW})

@inproceedings{kirillov2023segment,
  title={Segment anything},
  author={Kirillov, Alexander and Mintun, Eric and Ravi, Nikhila and Mao, Hanzi and Rolland, Chloe and Gustafson, Laura and Xiao, Tete and Whitehead, Spencer and Berg, Alexander C and Lo, Wan-Yen and others},
  booktitle={Proceedings of the IEEE/CVF international conference on computer vision},
  pages={4015--4026},
  year={2023}
}

@article{ravi2024sam,
  title={Sam 2: Segment anything in images and videos},
  author={Ravi, Nikhila and Gabeur, Valentin and Hu, Yuan-Ting and Hu, Ronghang and Ryali, Chaitanya and Ma, Tengyu and Khedr, Haitham and R{\"a}dle, Roman and Rolland, Chloe and Gustafson, Laura and others},
  journal={arXiv preprint arXiv:2408.00714},
  year={2024}
}

@inproceedings{liu2024grounding,
  title={Grounding {DINO}: Marrying {DINO} with grounded pre-training for open-set object detection},
  author={Liu, Shilong and Zeng, Zhaoyang and Ren, Tianhe and Li, Feng and Zhang, Hao and Yang, Jie and Jiang, Qing and Li, Chunyuan and Yang, Jianwei and Su, Hang and others},
  booktitle={European conference on computer vision},
  pages={38--55},
  year={2024},
  organization={Springer}
}

@inproceedings{yang2024depth,
  title={Depth anything: Unleashing the power of large-scale unlabeled data},
  author={Yang, Lihe and Kang, Bingyi and Huang, Zilong and Xu, Xiaogang and Feng, Jiashi and Zhao, Hengshuang},
  booktitle={Proceedings of the IEEE/CVF conference on computer vision and pattern recognition},
  pages={10371--10381},
  year={2024}
}

@article{yang2024depth2,
  title={Depth anything v2},
  author={Yang, Lihe and Kang, Bingyi and Huang, Zilong and Zhao, Zhen and Xu, Xiaogang and Feng, Jiashi and Zhao, Hengshuang},
  journal={Advances in Neural Information Processing Systems},
  volume={37},
  pages={21875--21911},
  year={2024}
}

@article{mildenhall2021nerf,
  title={{NeRF}: Representing scenes as neural radiance fields for view synthesis},
  author={Mildenhall, Ben and Srinivasan, Pratul P and Tancik, Matthew and Barron, Jonathan T and Ramamoorthi, Ravi and Ng, Ren},
  journal={Communications of the ACM},
  volume={65},
  number={1},
  pages={99--106},
  year={2021},
  publisher={ACM New York, NY, USA}
}

@article{kerbl20233d,
  title={3{D} {G}aussian splatting for real-time radiance field rendering.},
  author={Kerbl, Bernhard and Kopanas, Georgios and Leimk{\"u}hler, Thomas and Drettakis, George},
  journal={ACM Trans. Graph.},
  volume={42},
  number={4},
  pages={139--1},
  year={2023}
}

@incollection{levoy2023light,
  title={Light field rendering},
  author={Levoy, Marc and Hanrahan, Pat},
  booktitle={Seminal Graphics Papers: Pushing the Boundaries, Volume 2},
  pages={441--452},
  year={2023}
}

@phdthesis{ng2005light,
  title={Light field photography with a hand-held plenoptic camera},
  author={Ng, Ren and Levoy, Marc and Br{\'e}dif, Mathieu and Duval, Gene and Horowitz, Mark and Hanrahan, Pat},
  year={2005},
  school={Stanford university}
}

@article{bergen1991plenoptic,
  title={The plenoptic function and the elements of early vision},
  author={Bergen, James R and Adelson, Edward H},
  journal={Computational models of visual processing},
  volume={1},
  number={8},
  pages={3},
  year={1991}
}

@article{wang2022disentangling,
  title={Disentangling light fields for super-resolution and disparity estimation},
  author={Wang, Yingqian and Wang, Longguang and Wu, Gaochang and Yang, Jungang and An, Wei and Yu, Jingyi and Guo, Yulan},
  journal={IEEE Transactions on Pattern Analysis and Machine Intelligence},
  volume={45},
  number={1},
  pages={425--443},
  year={2022},
  publisher={IEEE}
}

@article{chao2023learning,
  title={Learning sub-pixel disparity distribution for light field depth estimation},
  author={Chao, Wentao and Wang, Xuechun and Wang, Yingqian and Wang, Guanghui and Duan, Fuqing},
  journal={IEEE Transactions on Computational Imaging},
  volume={9},
  pages={1126--1138},
  year={2023},
  publisher={IEEE}
}

@article{wanner2013variational,
  title={Variational light field analysis for disparity estimation and super-resolution},
  author={Wanner, Sven and Goldluecke, Bastian},
  journal={IEEE transactions on pattern analysis and machine intelligence},
  volume={36},
  number={3},
  pages={606--619},
  year={2013},
  publisher={IEEE}
}

@inproceedings{jeon2015accurate,
  title={Accurate depth map estimation from a lenslet light field camera},
  author={Jeon, Hae-Gon and Park, Jaesik and Choe, Gyeongmin and Park, Jinsun and Bok, Yunsu and Tai, Yu-Wing and So Kweon, In},
  booktitle={Proceedings of the IEEE conference on computer vision and pattern recognition},
  pages={1547--1555},
  year={2015}
}

@inproceedings{wen2024foundationpose,
  title={{FoundationPose}: Unified {6D} pose estimation and tracking of novel objects},
  author={Wen, Bowen and Yang, Wei and Kautz, Jan and Birchfield, Stan},
  booktitle={Proceedings of the IEEE/CVF Conference on Computer Vision and Pattern Recognition},
  pages={17868--17879},
  year={2024}
}

@inproceedings{zhang2022ssp,
  title={{SSP-Pose}: Symmetry-aware shape prior deformation for direct category-level object pose estimation},
  author={Zhang, Ruida and Di, Yan and Manhardt, Fabian and Tombari, Federico and Ji, Xiangyang},
  booktitle={2022 IEEE/RSJ International Conference on Intelligent Robots and Systems (IROS)},
  pages={7452--7459},
  year={2022},
  organization={IEEE}
}

@inproceedings{wang2019normalized,
  title={Normalized object coordinate space for category-level {6D} object pose and size estimation},
  author={Wang, He and Sridhar, Srinath and Huang, Jingwei and Valentin, Julien and Song, Shuran and Guibas, Leonidas J},
  booktitle={Proceedings of the IEEE/CVF conference on computer vision and pattern recognition},
  pages={2642--2651},
  year={2019}
}

@inproceedings{zheng2023hs,
  title={{HS-Pose}: Hybrid scope feature extraction for category-level object pose estimation},
  author={Zheng, Linfang and Wang, Chen and Sun, Yinghan and Dasgupta, Esha and Chen, Hua and Leonardis, Ale{\v{s}} and Zhang, Wei and Chang, Hyung Jin},
  booktitle={Proceedings of the IEEE/CVF conference on computer vision and pattern recognition},
  pages={17163--17173},
  year={2023}
}

@inproceedings{park2020latentfusion,
  title={Latentfusion: End-to-end differentiable reconstruction and rendering for unseen object pose estimation},
  author={Park, Keunhong and Mousavian, Arsalan and Xiang, Yu and Fox, Dieter},
  booktitle={Proceedings of the IEEE/CVF conference on computer vision and pattern recognition},
  pages={10710--10719},
  year={2020}
}

@inproceedings{rusu2009fast,
  title={Fast point feature histograms {(FPFH)} for 3D registration},
  author={Rusu, Radu Bogdan and Blodow, Nico and Beetz, Michael},
  booktitle={2009 IEEE international conference on robotics and automation},
  pages={3212--3217},
  year={2009},
  organization={IEEE}
}

@article{kingma2014adam,
  title={Adam: A method for stochastic optimization},
  author={Kingma, Diederik P},
  journal={arXiv preprint arXiv:1412.6980},
  year={2014}
}

@inproceedings{goncharov2025segment,
  title={Segment Anything in Light Fields for Real-Time Applications via Constrained Prompting},
  author={Goncharov, Nikolai and Dansereau, Donald},
  booktitle={Proceedings of the Winter Conference on Applications of Computer Vision},
  pages={1490--1496},
  year={2025}
}

@inproceedings{sun2022onepose,
  title={{OnePose}: One-shot object pose estimation without cad models},
  author={Sun, Jiaming and Wang, Zihao and Zhang, Siyu and He, Xingyi and Zhao, Hongcheng and Zhang, Guofeng and Zhou, Xiaowei},
  booktitle={Proceedings of the IEEE/CVF Conference on Computer Vision and Pattern Recognition},
  pages={6825--6834},
  year={2022}
}

@inproceedings{wiersma2025uncertainty,
  title={Uncertainty for {SVBRDF} Acquisition using Frequency Analysis},
  author={Wiersma, Ruben and Philip, Julien and Ha{\v{s}}an, Milo{\v{s}} and Mullia, Krishna and Luan, Fujun and Eisemann, Elmar and Deschaintre, Valentin},
  booktitle={Proceedings of the Special Interest Group on Computer Graphics and Interactive Techniques Conference Conference Papers},
  pages={1--12},
  year={2025}
}

@inproceedings{park2017coloredicp,
  title={Colored point cloud registration revisited},
  author={Park, Jaesik and Zhou, Qian-Yi and Koltun, Vladlen},
  booktitle={Proceedings of the IEEE International Conference on Computer Vision (ICCV)},
  year={2017},
  pages={143--152}
}

@article{wu_light_2017,
	title = {Light {Field} {Image} {Processing}: {An} {Overview}},
	volume = {11},
	issn = {1941-0484},
	shorttitle = {Light {Field} {Image} {Processing}},
	url = {https://ieeexplore.ieee.org/abstract/document/8022901/figures},
	doi = {10.1109/JSTSP.2017.2747126},
	number = {7},
	urldate = {2025-09-10},
	journal = {IEEE Journal of Selected Topics in Signal Processing},
	author = {Wu, Gaochang and Masia, Belen and Jarabo, Adrian and Zhang, Yuchen and Wang, Liangyong and Dai, Qionghai and Chai, Tianyou and Liu, Yebin},
	month = oct,
	year = {2017},
	pages = {926--954},
}

@article{levoy_light_2006,
	title = {Light {Fields} and {Computational} {Imaging}},
	volume = {39},
	copyright = {https://ieeexplore.ieee.org/Xplorehelp/downloads/license-information/IEEE.html},
	issn = {0018-9162},
	url = {http://ieeexplore.ieee.org/document/1673328/},
	doi = {10.1109/MC.2006.270},
	language = {en},
	number = {8},
	urldate = {2025-01-24},
	journal = {Computer},
	author = {Levoy, M.},
	month = aug,
	year = {2006},
	pages = {46--55},
}

@article{zhou_review_2021,
	title = {Review of light field technologies},
	volume = {4},
	issn = {2524-4442},
	url = {https://doi.org/10.1186/s42492-021-00096-8},
	doi = {10.1186/s42492-021-00096-8},
	language = {en},
	number = {1},
	urldate = {2025-11-06},
	journal = {Visual Computing for Industry, Biomedicine, and Art},
	author = {Zhou, Shuyao and Zhu, Tianqian and Shi, Kanle and Li, Yazi and Zheng, Wen and Yong, Junhai},
	month = dec,
	year = {2021},
	pages = {29},
}

@article{wang_light_2026,
	title = {Light field collaborative perception for visual object tracking},
	volume = {171},
	issn = {0031-3203},
	url = {https://www.sciencedirect.com/science/article/pii/S003132032500857X},
	doi = {10.1016/j.patcog.2025.112196},
	urldate = {2025-11-06},
	journal = {Pattern Recognition},
	author = {Wang, Mianzhao and Shi, Fan and Cheng, Xu and Zhao, Meng},
	month = mar,
	year = {2026},
	pages = {112196},
}

@ARTICLE{wang2023_object,

  author={Wang, Mianzhao and Shi, Fan and Cheng, Xu and Zhao, Meng and Zhang, Yao and Jia, Chen and Tian, Weiwei and Chen, Shengyong},

  journal={IEEE Transactions on Industrial Informatics}, 

  title={Visual Object Tracking Based on Light-Field Imaging in the Presence of Similar Distractors}, 

  year={2023},

  volume={19},

  number={3},

  pages={2705-2716},

  doi={10.1109/TII.2022.3159648}}

@article{collet_moped_2011,
	title = {The {MOPED} framework: {Object} recognition and pose estimation for manipulation},
	volume = {30},
	issn = {0278-3649, 1741-3176},
	shorttitle = {The {MOPED} framework},
	url = {https://journals.sagepub.com/doi/10.1177/0278364911401765},
	doi = {10.1177/0278364911401765},
	language = {en},
	number = {10},
	urldate = {2025-11-06},
	journal = {The International Journal of Robotics Research},
	author = {Collet, Alvaro and Martinez, Manuel and Srinivasa, Siddhartha S},
	month = sep,
	year = {2011},
	pages = {1284--1306},
}

@article{loshchilov2017decoupled,
  title={Decoupled weight decay regularization},
  author={Loshchilov, Ilya and Hutter, Frank},
  journal={arXiv preprint arXiv:1711.05101},
  year={2017}
}

@article{ye2025gsplat,
  title={gsplat: An open-source library for Gaussian splatting},
  author={Ye, Vickie and Li, Ruilong and Kerr, Justin and Turkulainen, Matias and Yi, Brent and Pan, Zhuoyang and Seiskari, Otto and Ye, Jianbo and Hu, Jeffrey and Tancik, Matthew and others},
  journal={Journal of Machine Learning Research},
  volume={26},
  number={34},
  pages={1--17},
  year={2025}
}

@article{lepetit2005monocular,
  title={Monocular model-based {3D} tracking of rigid objects: A survey},
  author={Lepetit, Vincent and Fua, Pascal and others},
  journal={Foundations and Trends{\textregistered} in Computer Graphics and Vision},
  volume={1},
  number={1},
  pages={1--89},
  year={2005},
  publisher={Now Publishers, Inc.}
}

@article{
  nguyen2025gotrack,
  author    = {Nguyen, Van Nguyen and Forster, Christian and Tekin, Bugra and Shkodrani, Sindi and Lepetit, Vincent and Keskin, Cem and Hoda{\v{n}}, Tom{\'a}{\v{s}}},
  title     = {GoTrack: Generic {6DoF} Object Pose Refinement and Tracking},
  journal   = {Computer Vision and Patern Recognition Workshops (CVPRW)},
  year      = {2025},
}

@inproceedings{li2018deepim,
  title={Deepim: Deep iterative matching for {6D} pose estimation},
  author={Li, Yi and Wang, Gu and Ji, Xiangyang and Xiang, Yu and Fox, Dieter},
  booktitle={Proceedings of the European conference on computer vision (ECCV)},
  pages={683--698},
  year={2018}
}

@inproceedings{lin2022keypoint,
  title={Keypoint-based category-level object pose tracking from an RGB sequence with uncertainty estimation},
  author={Lin, Yunzhi and Tremblay, Jonathan and Tyree, Stephen and Vela, Patricio A and Birchfield, Stan},
  booktitle={2022 International Conference on Robotics and Automation (ICRA)},
  pages={1258--1264},
  year={2022},
  organization={IEEE}
}

@inproceedings{wen_bundletrack_2021,
	title = {{BundleTrack}: {6D} {Pose} {Tracking} for {Novel} {Objects} without {Instance} or {Category}-{Level} {3D} {Models}},
	shorttitle = {{BundleTrack}},
	url = {https://ieeexplore.ieee.org/abstract/document/9635991},
	doi = {10.1109/IROS51168.2021.9635991},
	urldate = {2025-11-11},
	booktitle = {2021 {IEEE}/{RSJ} {International} {Conference} on {Intelligent} {Robots} and {Systems} ({IROS})},
	author = {Wen, Bowen and Bekris, Kostas},
	month = sep,
	year = {2021},
	note = {ISSN: 2153-0866},
	pages = {8067--8074},
}

@inproceedings{wang20206,
  title={6-{PACK}: Category-level {6D} pose tracker with anchor-based keypoints},
  author={Wang, Chen and Mart{\'\i}n-Mart{\'\i}n, Roberto and Xu, Danfei and Lv, Jun and Lu, Cewu and Fei-Fei, Li and Savarese, Silvio and Zhu, Yuke},
  booktitle={2020 IEEE International Conference on Robotics and Automation (ICRA)},
  pages={10059--10066},
  year={2020},
  organization={IEEE}
}

@INPROCEEDINGS{RGB_D_Cai,
  author={Cai, Yuxiang and Zhu, Yifan and Zhang, Haiwei and Ren, Bo},
  booktitle={2023 IEEE/CVF International Conference on Computer Vision (ICCV)}, 
  title={Consistent Depth Prediction for Transparent Object Reconstruction from {RGB-D} Camera}, 
  year={2023},
  volume={},
  number={},
  pages={3436-3445},
  doi={10.1109/ICCV51070.2023.00320}}

@inproceedings{wen2023bundlesdf,
  title={{BundleSDF}: Neural {6-DOF} tracking and {3D} reconstruction of unknown objects},
  author={Wen, Bowen and Tremblay, Jonathan and Blukis, Valts and Tyree, Stephen and M{\"u}ller, Thomas and Evans, Alex and Fox, Dieter and Kautz, Jan and Birchfield, Stan},
  booktitle={Proceedings of the IEEE/CVF Conference on Computer Vision and Pattern Recognition},
  pages={606--617},
  year={2023}
}

@phdthesis{gray2025gradient,
  title={Gradient Consistency: A New Take on Variational Optical Flow and Disparity Estimation},
  author={Gray, James Lyndon},
  year={2025},
  school={University of New South Wales (Australia)}
}

@article{ma_differential_2020,
	title = {Differential {Scene} {Flow} from {Light} {Field} {Gradients}},
	volume = {128},
	issn = {1573-1405},
	url = {https://doi.org/10.1007/s11263-019-01230-z},
	doi = {10.1007/s11263-019-01230-z},
	language = {en},
	number = {3},
	urldate = {2025-01-24},
	journal = {International Journal of Computer Vision},
	author = {Ma, Sizhuo and Smith, Brandon M. and Gupta, Mohit},
	month = mar,
	year = {2020},
	pages = {679--697},
}

@article{gray2025multi,
  title={Multi-View Disparity Estimation Using the Gradient Consistency Model},
  author={Gray, James L and Naman, Aous T and Taubman, David S},
  journal={IEEE Transactions on Image Processing},
  year={2025},
  publisher={IEEE}
}

@inproceedings{tran_variational_2017,
	title = {Variational disparity estimation framework for plenoptic images},
	url = {https://ieeexplore.ieee.org/abstract/document/8019377},
	doi = {10.1109/ICME.2017.8019377},
	urldate = {2025-08-18},
	booktitle = {2017 {IEEE} {International} {Conference} on {Multimedia} and {Expo} ({ICME})},
	author = {Tran, Trung-Hieu and Wang, Zhe and Simon, Sven},
	month = jul,
	year = {2017},
	note = {ISSN: 1945-788X},
	pages = {1189--1194},
}

@inproceedings{tao2013depth,
  title={Depth from combining defocus and correspondence using light-field cameras},
  author={Tao, Michael W and Hadap, Sunil and Malik, Jitendra and Ramamoorthi, Ravi},
  booktitle={Proceedings of the IEEE international conference on computer vision},
  pages={673--680},
  year={2013}
}

@ARTICLE{OccCasNet_2024,
  author={Chao, Wentao and Duan, Fuqing and Wang, Xuechun and Wang, Yingqian and Lu, Ke and Wang, Guanghui},
  journal={IEEE Transactions on Computational Imaging}, 
  title={{OccCasNet}: Occlusion-Aware Cascade Cost Volume for Light Field Depth Estimation}, 
  year={2024},
  volume={10},
  number={},
  pages={1680-1691},
  doi={10.1109/TCI.2024.3488563}}

@InProceedings{Gao_2025_ICCV,
    author    = {Gao, Chen and Zhang, Shuo and Lin, Youfang},
    title     = {Epipolar Consistent Attention Aggregation Network for Unsupervised Light Field Disparity Estimation},
    booktitle = {Proceedings of the IEEE/CVF International Conference on Computer Vision (ICCV)},
    month     = {October},
    year      = {2025},
    pages     = {6488-6497}
}

@article{wang_unisdf_2024,
	title = {{UniSDF}: {Unifying} {Neural} {Representations} for {High}-{Fidelity} {3D} {Reconstruction} of {Complex} {Scenes} with {Reflections}},
	volume = {37},
	shorttitle = {{UniSDF}},
	url = {https://proceedings.neurips.cc/paper_files/paper/2024/hash/05b12f103c9e613efc4c85674cdc9066-Abstract-Conference.html},
	language = {en},
	urldate = {2025-09-24},
	journal = {Advances in Neural Information Processing Systems},
	author = {Wang, Fangjinhua and Rakotosaona, Marie-Julie and Niemeyer, Michael and Szeliski, Richard and Pollefeys, Marc and Tombari, Federico},
	month = dec,
	year = {2024},
	pages = {3157--3184},
}

@misc{verbin_ref-nerf_2021,
	title = {Ref-{NeRF}: {Structured} {View}-{Dependent} {Appearance} for {Neural} {Radiance} {Fields}},
	shorttitle = {Ref-{NeRF}},
	url = {http://arxiv.org/abs/2112.03907},
	doi = {10.48550/arXiv.2112.03907},
	urldate = {2025-09-29},
	publisher = {arXiv},
	author = {Verbin, Dor and Hedman, Peter and Mildenhall, Ben and Zickler, Todd and Barron, Jonathan T. and Srinivasan, Pratul P.},
	month = dec,
	year = {2021},
	note = {arXiv:2112.03907 [cs]},
}

@inproceedings{gu_irgs_2025,
	title = {{IRGS}: {Inter}-{Reflective} {Gaussian} {Splatting} with {2D} {Gaussian} {Ray} {Tracing}},
	shorttitle = {{IRGS}},
    booktitle = {Proceedings of the IEEE/CVF Conference on Computer Vision and Pattern Recognition (CVPR)},
	url = {https://openaccess.thecvf.com/content/CVPR2025/html/Gu_IRGS_Inter-Reflective_Gaussian_Splatting_with_2D_Gaussian_Ray_Tracing_CVPR_2025_paper.html},
	language = {en},
	urldate = {2025-08-28},
	author = {Gu, Chun and Wei, Xiaofei and Zeng, Zixuan and Yao, Yuxuan and Zhang, Li},
	year = {2025},
	pages = {10943--10952},
}

@inproceedings{ye_3d_2024,
	address = {New York, NY, USA},
	series = {{SIGGRAPH} '24},
	title = {{3D} {Gaussian} {Splatting} with {Deferred} {Reflection}},
	isbn = {979-8-4007-0525-0},
	url = {https://dl.acm.org/doi/10.1145/3641519.3657456},
	doi = {10.1145/3641519.3657456},
	urldate = {2024-12-17},
	booktitle = {{ACM} {SIGGRAPH} 2024 {Conference} {Papers}},
	publisher = {Association for Computing Machinery},
	author = {Ye, Keyang and Hou, Qiming and Zhou, Kun},
	month = jul,
	year = {2024},
	pages = {1--10},
}

@misc{chen_gsgtrack_2024,
	title = {{GSGTrack}: {Gaussian} {Splatting}-{Guided} {Object} {Pose} {Tracking} from {RGB} {Videos}},
	shorttitle = {{GSGTrack}},
	url = {http://arxiv.org/abs/2412.02267},
	doi = {10.48550/arXiv.2412.02267},
	language = {en},
	urldate = {2025-11-12},
	publisher = {arXiv},
	author = {Chen, Zhiyuan and Lu, Fan and Yu, Guo and Li, Bin and Qu, Sanqing and Huang, Yuan and Fu, Changhong and Chen, Guang},
	month = dec,
	year = {2024},
	note = {arXiv:2412.02267 [cs]},
}

@misc{jin_6dope-gs_2025,
	title = {{6DOPE}-{GS}: {Online} {6D} {Object} {Pose} {Estimation} using {Gaussian} {Splatting}},
	shorttitle = {{6DOPE}-{GS}},
	url = {http://arxiv.org/abs/2412.01543},
	doi = {10.48550/arXiv.2412.01543},
	language = {en},
	urldate = {2025-11-12},
	publisher = {arXiv},
	author = {Jin, Yufeng and Prasad, Vignesh and Jauhri, Snehal and Franzius, Mathias and Chalvatzaki, Georgia},
	month = apr,
	year = {2025},
	note = {arXiv:2412.01543 [cs]},
}

@Manual{Blender,
   title = {Blender - a 3D modelling and rendering package},
   author = {Blender Online Community},
   organization = {Blender Foundation},
   address = {Stichting Blender Foundation, Amsterdam},
   year = {2018},
   url = {http://www.blender.org},
 }

@article{awais_foundation_2025,
	title = {Foundation {Models} {Defining} a {New} {Era} in {Vision}: {A} {Survey} and {Outlook}},
	volume = {47},
	issn = {1939-3539},
	shorttitle = {Foundation {Models} {Defining} a {New} {Era} in {Vision}},
	url = {https://ieeexplore.ieee.org/document/10834497},
	doi = {10.1109/TPAMI.2024.3506283},
	number = {4},
	urldate = {2025-11-12},
	journal = {IEEE Transactions on Pattern Analysis and Machine Intelligence},
	author = {Awais, Muhammad and Naseer, Muzammal and Khan, Salman and Anwer, Rao Muhammad and Cholakkal, Hisham and Shah, Mubarak and Yang, Ming-Hsuan and Khan, Fahad Shahbaz},
	month = apr,
	year = {2025},
	pages = {2245--2264},
}

@InProceedings{Chen_2025_CVPR,
    author    = {Chen, Sili and Guo, Hengkai and Zhu, Shengnan and Zhang, Feihu and Huang, Zilong and Feng, Jiashi and Kang, Bingyi},
    title     = {Video Depth Anything: Consistent Depth Estimation for Super-Long Videos},
    booktitle = {Proceedings of the IEEE/CVF Conference on Computer Vision and Pattern Recognition (CVPR)},
    month     = {June},
    year      = {2025},
    pages     = {22831-22840}
}

@misc{xiang_posecnn_2017,
	title = {{PoseCNN}: {A} {Convolutional} {Neural} {Network} for {6D} {Object} {Pose} {Estimation} in {Cluttered} {Scenes}},
	shorttitle = {{PoseCNN}},
	url = {https://arxiv.org/abs/1711.00199v3},
	language = {en},
	urldate = {2025-11-14},
	journal = {arXiv.org},
	author = {Xiang, Yu and Schmidt, Tanner and Narayanan, Venkatraman and Fox, Dieter},
	month = nov,
	year = {2017},
}
}

\clearpage
\setcounter{page}{1}
\maketitlesupplementary

\subsection{Datasets}
\begin{figure}
    \centering
    \includegraphics[width=0.8\linewidth]{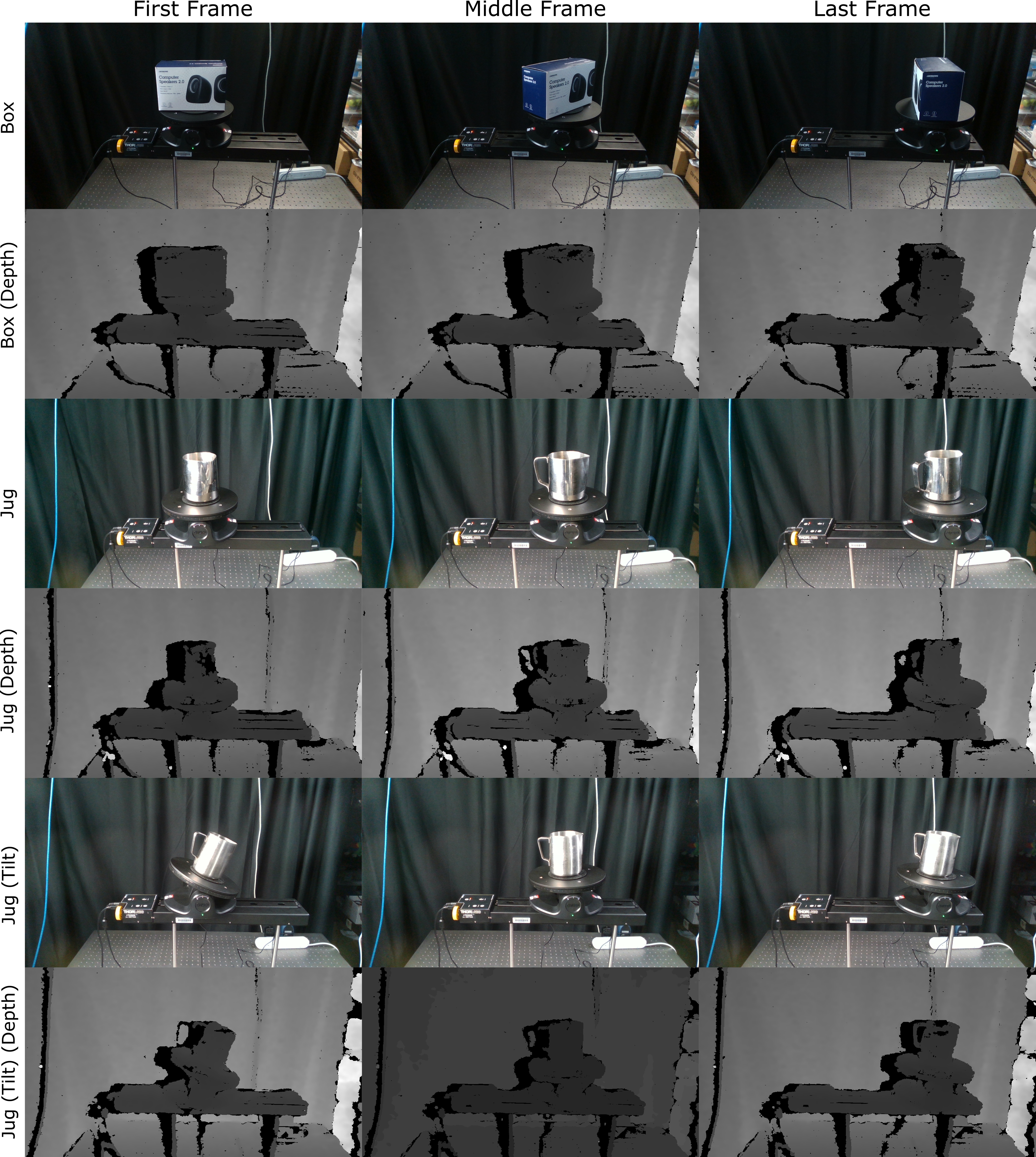}
    \caption{The first, middle and final frames of the Box, Jug and Jug (Tilt) sequences with their depth maps.}
    \label{fig:Dataset_1}
\end{figure}

\begin{figure}
    \centering
    \includegraphics[width=0.8\linewidth]{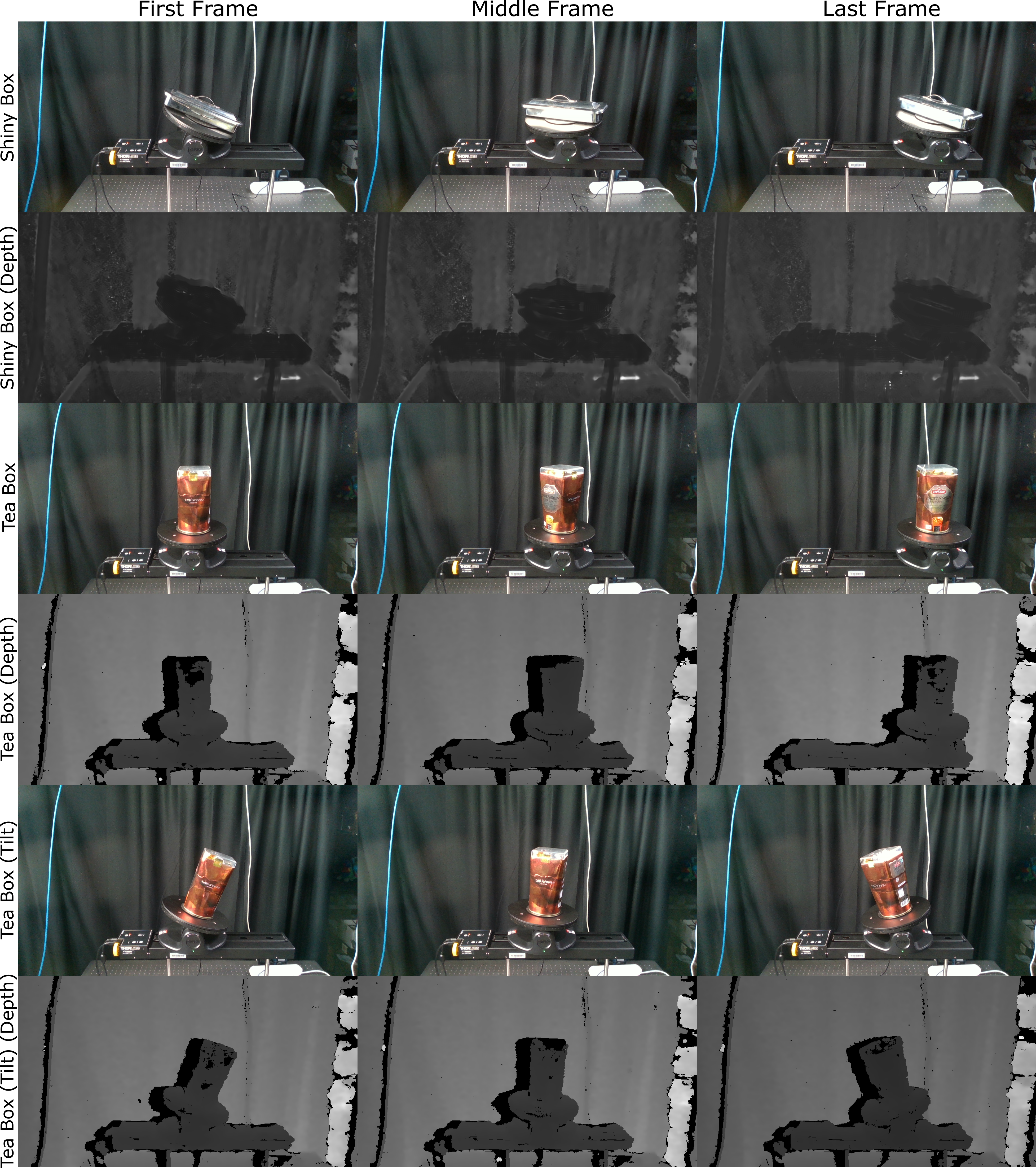}
    \caption{The first, middle and final frames of the Shiny Box, Tea Box and Tea Box (Tilt) sequences with their depth maps.}
    \label{fig:Dataset_2}
\end{figure}

\begin{figure}
    \centering
    \includegraphics[width=0.8\linewidth]{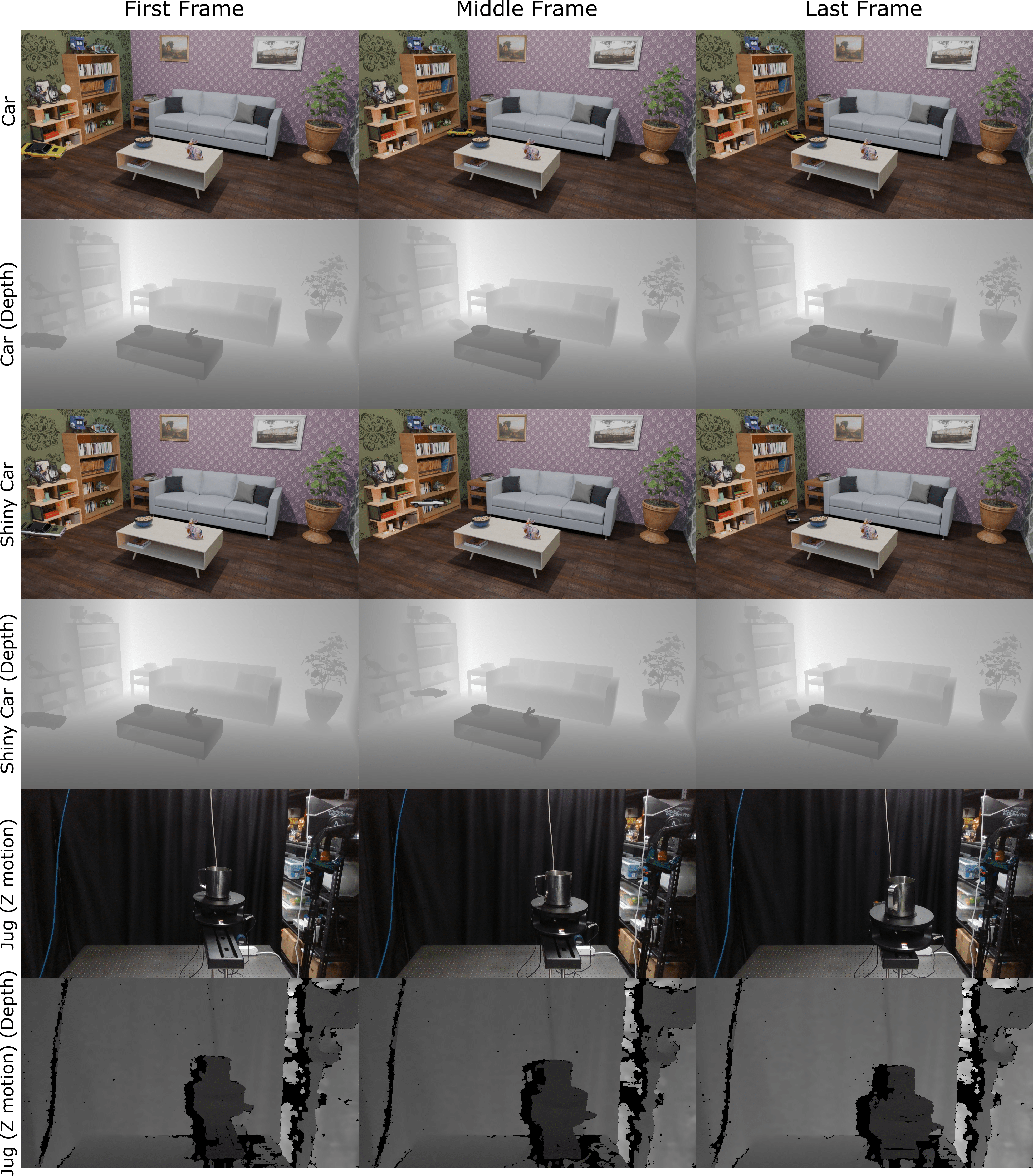}
    \caption{The first, middle and final frames of the Jug (Z motion), Car, Shiny Car synthetic sequences with their depth maps.}
    \label{fig:Dataset_3}
\end{figure}

\begin{figure*}
    \centering
    \includegraphics[width=\linewidth]{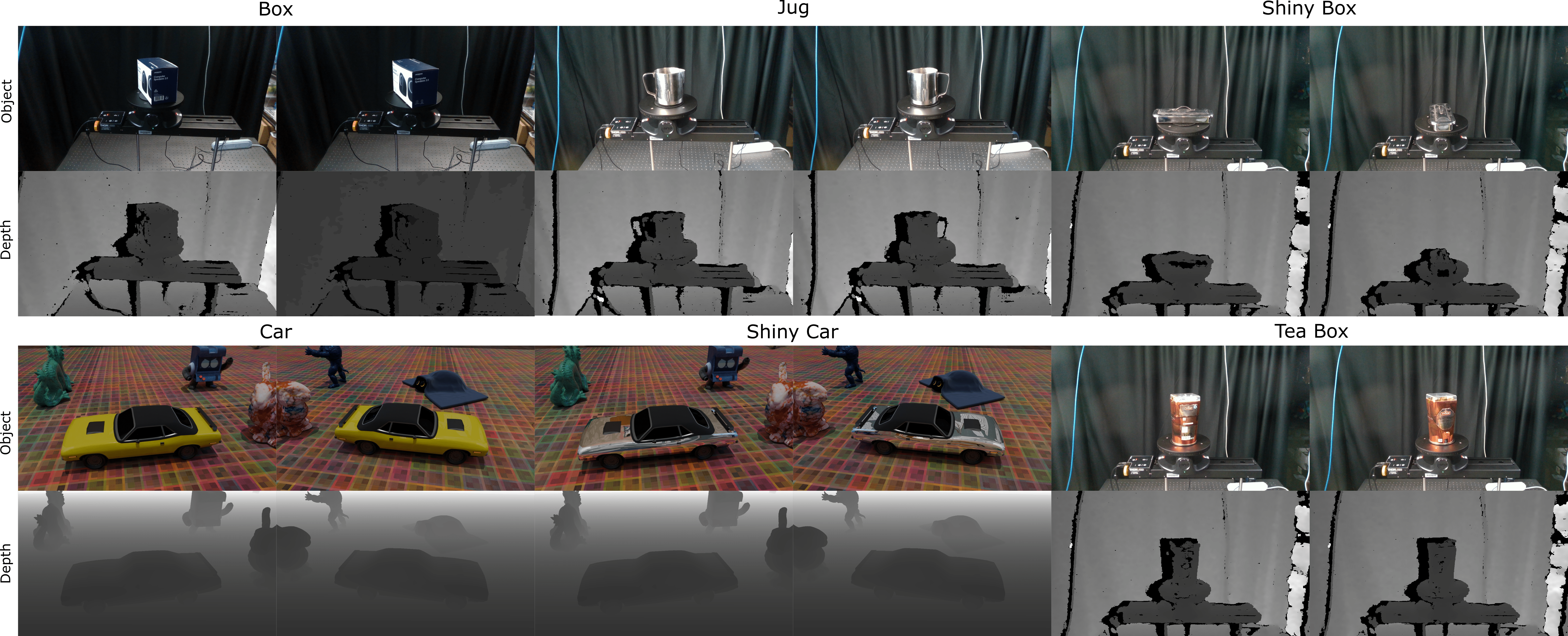}
    \caption{Front and back/side views from the reference sequences for each object. Side views instead of back views are included for the Box and Shiny Box objects since these objects are symmetric.}
    \label{fig:Ref_1}
\end{figure*}

As discussed in the introduction, we present a light field pose tracking dataset. 
We present the first, middle, and last frames of each sequence in the dataset in Figures~\ref{fig:Dataset_1}, \ref{fig:Dataset_2} and \ref{fig:Dataset_3}. 
This dataset also contains reference sequences that can be used to initialise methods such as FoundationPose prior to object tracking. 
We show two frames from each reference sequence in Figure~\ref{fig:Ref_1} depicting a front view and a side or back view depending on the object's symmetry.


\subsection{Implementation Details}
\label{sec:impl_details}
\textbf{Training of the Gaussians} \quad
We optimize Gaussian centroids, view-dependent color, scales, opacities and quaternions using AdamW \cite{loshchilov2017decoupled} with base learning rate of $10^{-3}$,  $\beta = (0.9, 0.99)$, and 1000 epochs, with opacity learning rate by reduced $0.5$. We progressively enable higher–order spherical harmonic coefficients during training, starting by optimizing only the geometry (means, scales, orientations, opacities), and every 50 epochs unfreezing the next spherical harmonic coefficient up to degree 3. To prevent overfitting to a particular view, each epoch all quaternions are normalized to unit length, and scales are clamped to be within $[d_{\mathrm{scene}} \cdot 10^{-3},  d_{\mathrm{scene}} \cdot 10^{-2}]$, where $d_{\mathrm{scene}}$ is the scene diagonal. 
\\
\textbf{Coarse Registration} \quad
For both input point clouds, we first applied voxel-grid downsampling with a voxel size of 0.02 m. Surface normals were then estimated using a spherical neighborhood of radius 0.04 m capped at 30 nearest neighbors. On the downsampled geometry, we computed fast point feature histograms \cite{rusu2009fast} descriptors within a radius of 0.1 m using up to 100 neighbors. Global alignment was obtained via RANSAC-based feature matching, using $n_{\mathrm{corresp}} = 4$ and a maximum correspondence threshold of 0.05 m. The resulting pose served as initialization for the refinement stage, where we applied colored ICP \cite{park2017coloredicp} on the full-resolution point clouds with a maximum correspondence distance of 0.02 m.

\textbf{Pose Refinement} \quad
We refine the relative transform between the two sets of Gaussians by jointly optimizing translation, rotation, ambient color coefficients, scales, opacities and quaternions. We use base learning rate of $10^{-3}$, while reducing it by $0.1 \times$ for translation, $0.5 \times$ for opacity, and $0.1 \times$ for ambient color, scales and quaternions. We reduce learning rate for each parameter by 0.1 every 20 steps, training a total of 400 steps. Instead of relying on the final optimization step, we select the relative transform corresponding to the minimum loss value observed throughout all epochs. While calculating the photometric loss, we turn off the view-dependent spherical harmonics for rasterization. As a result, our Gaussians are trained to provide pure ambient color that is clearly decoupled from the specular characteristics that interfere with pose estimation.

\end{document}